\documentclass[utf8]{FrontiersinHarvard} 
\usepackage{url,hyperref,lineno,microtype,subcaption}
\usepackage[onehalfspacing]{setspace}

\def\keyFont{\fontsize{8}{11}\helveticabold }
\def\firstAuthorLast{Gabriel {et~al.}} 
\def\Authors{
    Paolo Gabriel\,$^{1,*}$, 
    Peter Rehani\,$^{1}$, 
    Tyler Troy\,$^{1}$, 
    Tiffany Wyatt\,$^{1}$,
    Michael Choma\,$^{1}$, 
    and Narinder Singh\,$^{1}$
}
\def\Address{$^{1}$LookDeep Health, Oakland, CA, USA}
\def\corrAuthor{Paolo Gabriel}

\def\corrEmail{paolo@lookdeep.health}

\begin{document}
\onecolumn
\firstpage{1}

\title[Continuous Patient Monitoring with Vision AI]{Continuous Patient Monitoring with AI: Real-Time Analysis of Video in Hospital Care Settings} 

\author[\firstAuthorLast ]{\Authors}
\address{} 
\correspondance{} 
\extraAuth{}

\maketitle

\begin{abstract}

\section{}
This study introduces an AI-driven platform for continuous and passive patient monitoring in hospital settings, developed by LookDeep Health. 
Leveraging advanced computer vision, the platform provides real-time insights into patient behavior and interactions through video analysis, securely storing inference results in the cloud for retrospective evaluation. 
The dataset, compiled in collaboration with 11 hospital partners, encompasses over 300 high-risk fall patients and over 1,000 days of inference, enabling applications such as fall detection and safety monitoring for vulnerable patient populations. 
To foster innovation and reproducibility, an anonymized subset of this dataset is publicly available. 
The AI system detects key components in hospital rooms, including individuals' presence and roles, furniture location, motion magnitude, and boundary crossings.
Performance evaluation demonstrates strong accuracy in object detection (macro F1-score = 0.92) and patient-role classification (F1-score = 0.98), as well as reliable trend analysis for the “patient alone” metric (mean logistic regression accuracy = 0.82 ± 0.15). 
These capabilities enable automated detection of patient isolation, wandering, or unsupervised movement—key indicators for fall risk and other adverse events. 
This work establishes benchmarks for validating AI-driven patient monitoring systems, highlighting the platform's potential to enhance patient safety and care by providing continuous, data-driven insights into patient behavior and interactions.

\tiny
 \keyFont{ \section{Keywords:} Artificial Intelligence, Medical Imaging, Computer Vision, Patient Monitoring, RGB Video, Deep Learning, Healthcare Analytics}
\end{abstract}

\section{Introduction}
In hospitals, direct patient observation is limited—nurses spend only 37\% of their shift engaged in patient care (\cite{westbrook2011much}), and physicians average just 10 visits per hospital stay (\cite{chae2021improved}).
This limited interaction hinders the ability to fully understand patient behaviors, such as how often patients are left alone, how much they move unsupervised, and how care allocation varies by time or condition.
Virtual monitoring systems, which allow remote patient observation via audio-video devices, have improved safety, particularly for high-risk patients (\cite{abbe2021continuous}).

Artificial Intelligence (AI) is transforming healthcare by enhancing diagnostic accuracy, streamlining data processing, and personalizing patient care (\cite{davenport2019potential}, \cite{davoudi2019intelligent}, \cite{bajwa2021artificial}).
While AI has found success in tasks like surgical assistance (\cite{mascagni2022computer}) and diagnostic imaging (\cite{esteva2021deep}), patient monitoring represents a critical frontier.
Unlike these tasks, continuous patient monitoring involves real-time video analysis over extended periods, requiring AI systems to process data efficiently and extract actionable insights spanning days, like day-over-day movement (\cite{parker2022continuous}).

Continuous monitoring enhances safety and enables the detection of risks often missed during periodic assessments. 
For example, trends like delirium fluctuate throughout the day, but infrequent observations make these patterns hard to capture (\cite{wilson2020delirium}). 
Similarly, patients occasionally leave their beds unattended—a key fall risk—yet monitoring every instance in real-time remains challenging. 
A robust computer vision-based system can provide immediate, context-aware insights into patient behavior (\cite{chen2018patient}), caregiver interactions (\cite{avogaro2023markerless}), and room conditions (\cite{haque2020illuminating}).
Such systems surpass traditional intermittent observation methods by detecting subtle patterns that inform care decisions (\cite{lindroth2024applied}). 

However, achieving scalability, transparency, and adaptability in continuous monitoring systems presents significant challenges. 
These include efficiently processing video data at higher frame-rates (\cite{posch2014retinomorphic}), ensuring privacy compliance (\cite{watzlaf2010voip}), and adapting to dynamic hospital settings with varying lighting, camera angles, and patient behaviors. 
Addressing these technical and operational challenges is critical for AI-driven monitoring systems to gain acceptance and deliver meaningful outcomes, such as reducing falls and other preventable harms.

\begin{figure}[h!]
\begin{center}
\includegraphics[width=12cm]{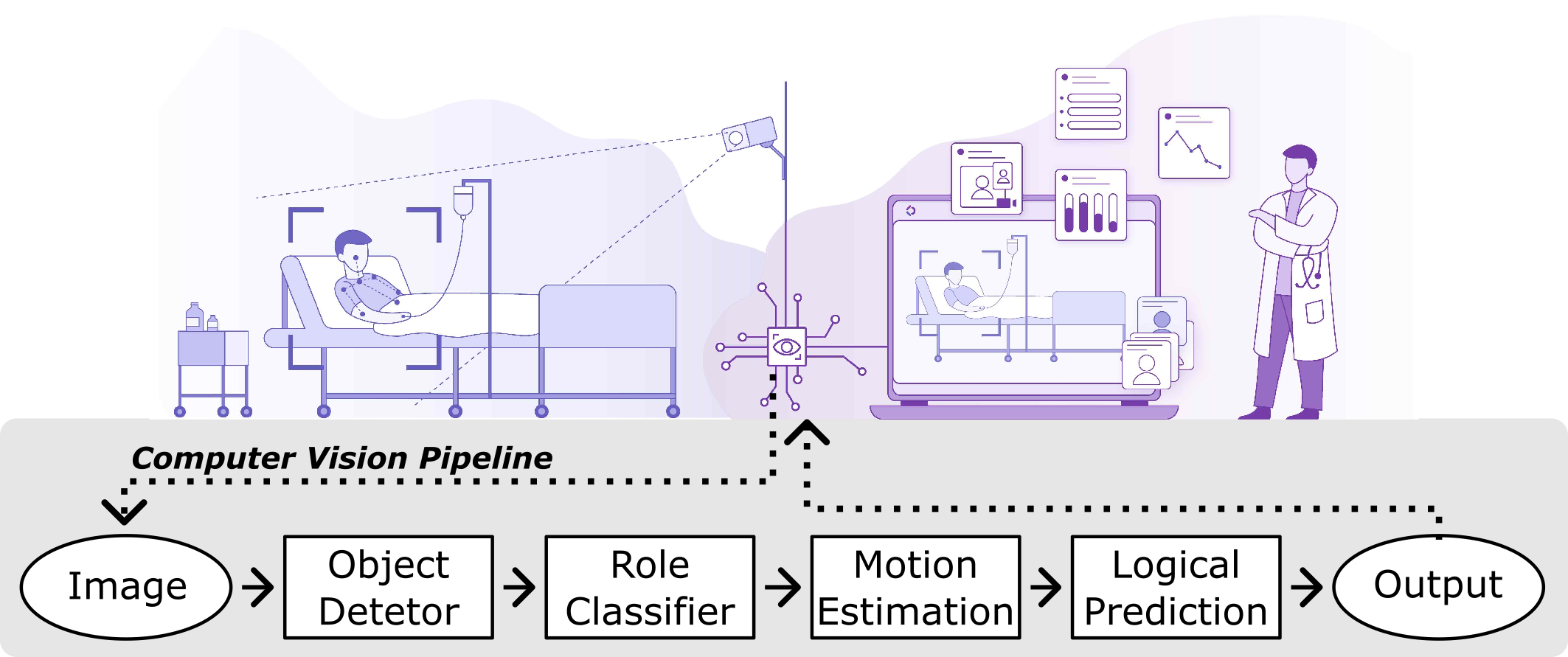}
\end{center}
\caption{\textbf{Illustrative Workflow of the LookDeep Health AI-driven Patient Monitoring Platform.}
The system captures video from a hospital room using mounted cameras and processes each image through a series of computer vision modules. 
The output is presented as real-time insights for healthcare staff.
}\label{fig:overview}
\end{figure}

To bridge these gaps, this research presents a novel AI-driven system for continuous patient monitoring using RGB video (Figure \ref{fig:overview}), developed collaboratively with industry and healthcare providers.
The LookDeep Health platform aims to enhance patient care by providing real-time monitoring and producing computer-vision-based insights into patient behavior, movement, and interactions with healthcare staff.

This study offers several key contributions:
\begin{enumerate}
    \item \textbf{Implementation of Advanced Computer Vision Models:} Our system utilizes state-of-the-art models for real-time predictions, including localization of people and furniture, monitoring boundary crossings, and calculating motion scores.
    \item \textbf{Real-World Validation:} We rigorously evaluated the system's performance in live hospital settings, illustrating its capability to present care providers with accurate data from continuous monitoring, and laying the foundation for future AI-enabled patient monitoring solutions.
    \item \textbf{Dataset Development:} We developed a comprehensive dataset encompassing over 300 high-risk fall patients tracked across 1,000 collective days and 11 hospitals, creating a valuable resource for studying patient behavior and hospital care patterns. This dataset is publicly available for further research at \href{https://github.com/lookdeep/ai-norms-2024}{lookdeep/ai-norms-2024} 
\end{enumerate}

\section{Methods}
\subsection{Study Design}
The LookDeep Health patient monitoring platform was deployed across 11 hospitals in three states within a single healthcare network. 
The system provides continuous, real-time monitoring of high-risk fall patients.
Data collection adhered to institutional guidelines and patient consent procedures (see \textit{Research Ethics}).

\subsubsection{Participants}
Patients monitored by LookDeep Health were primarily high-risk fall patients identified through mobility assessments as part of standard care protocols. 
This classification often results in the patient also being categorized as non-ambulatory during the inpatient stay (\cite{capo2023revealing}).

\begin{figure}[h!]
    \begin{center}
    \includegraphics[width=14cm]{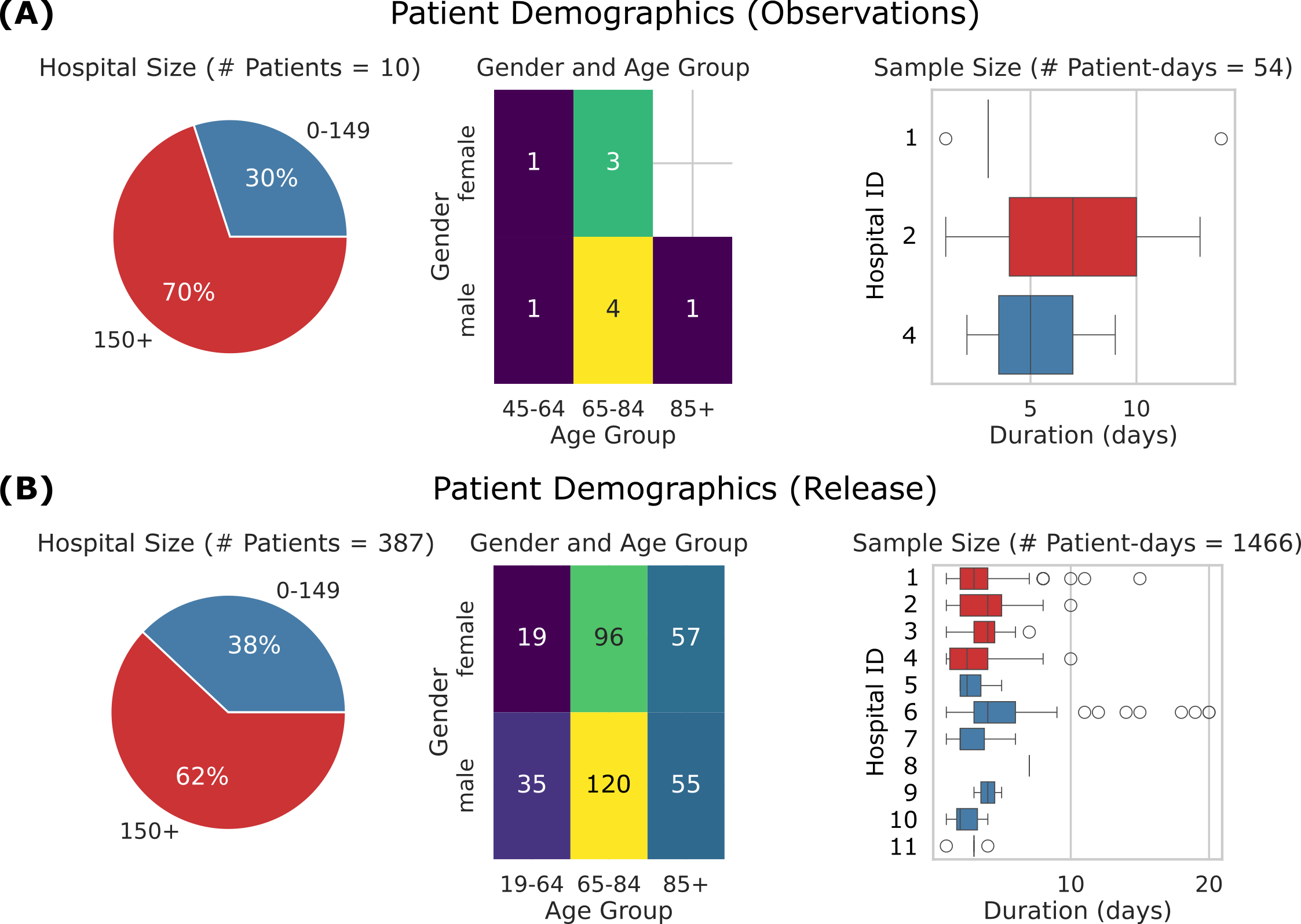}
    \end{center}
    \caption{\textbf{Overview of Patient Demographics.}
    (A) Observations subset, comprising 10 patients, 54 patient-days, and 3 hospitals.
    (B) Released dataset, comprising 387 patients, 1,466 patient-days, and 11 hospitals.
    Left: Pie charts showing distribution of hospitals by size. 
    Hospitals are grouped by average daily census. 
    Center: Heat maps showing patient age distribution by gender.
    Right: Box plots showing patient length of monitoring.
    Central line represents the median, box edges indicate the 25th and 75th percentiles, and whiskers extend to the most extreme data points within 1.5 times the interquartile range.
    The points represent outliers beyond this range.
    The released dataset shows a broader demographic and extended data duration compared to the observations subset.
    }\label{fig:demographic}
\end{figure}

Data was organized into three subsets:
\begin{enumerate}
    \item \textbf{Single-Frame Analysis:} Periodic samples from monitoring sessions were used for training and testing object detectors, with over 40,000 frames collected to date. Only patients monitored during the first week of each month were included in the test set, providing 10,000 frames held out for consistent model evaluation. 
    \item \textbf{Observation Logging:} Ten patients who experienced falls were selected for additional annotation over a twelve month period (Figure \ref{fig:demographic}A).
    \item \textbf{Public Dataset:} Over 300 high-risk fall patients were monitored during a six month period, excluding those monitored for less than two days (Figure \ref{fig:demographic}B).
\end{enumerate}

\begin{figure}[h!]
    \begin{center}
    \includegraphics[width=8cm]{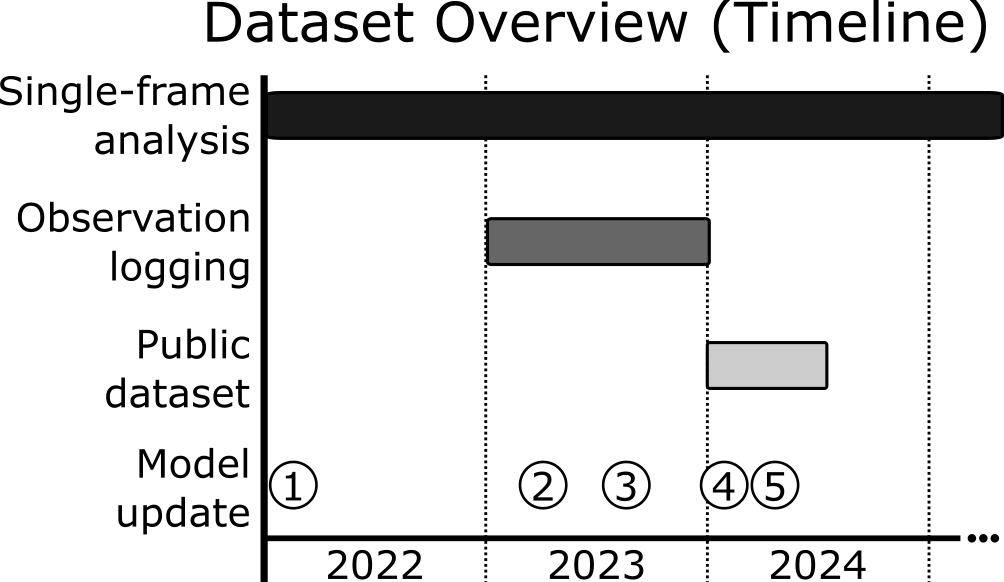}
    \end{center}
    \caption{\textbf{Dataset Overview and Timeline of Model Updates.} 
    Progression of data collection and model updates for the LookDeep Health monitoring system. 
    Single-frame analysis data collection spans a two year period - a broad temporal range for training and validation of object detection and classification tasks. 
    Observation logging data, used for trend validation, was collected over a one year period. 
    The publicly released dataset includes data from a more recent six month period, representing over 1,000 collective patient days. 
    Model updates are indicated by numbered points.
    }\label{fig:timeline}
\end{figure}

As shown in Figure \ref{fig:timeline}, data collection spanned multiple years, with each subset contributing to the development and validation of the AI system, with some overlap between subsets.

\subsubsection{Patient Monitoring System Overview}
\begin{figure}[h!]
    \begin{center}
    \includegraphics[width=12cm]{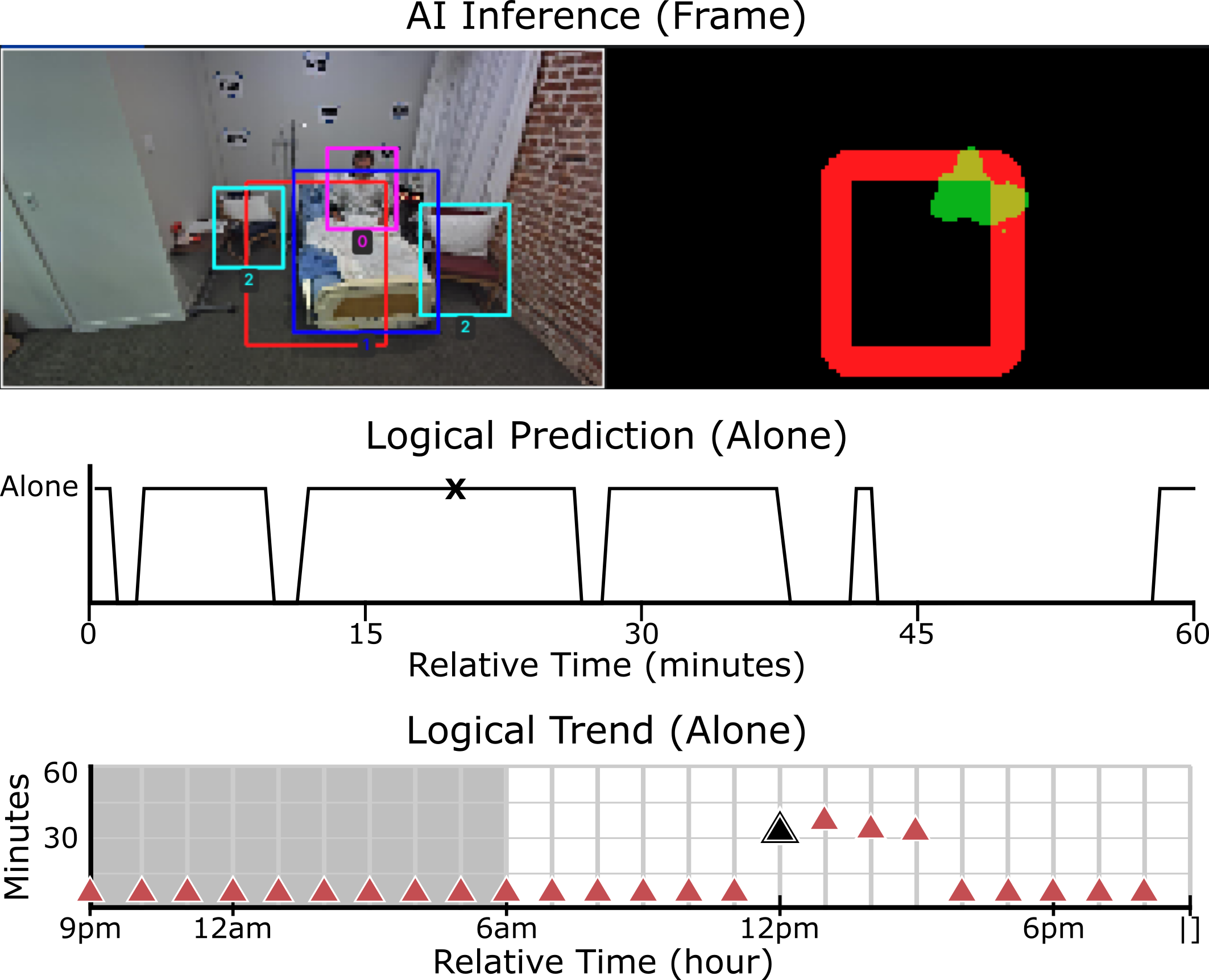}
    \end{center}
    \caption{\textbf{Real-Time Object Detection, Motion Analysis, and Patient Status Monitoring.} 
    Top-left: Object detection with bounding boxes.
    Top-right: Segmentation map (red = safety zone, green = motion).
    Middle: "Alone" logical trend over time for every second within the hour.
    Bottom: "Alone" trend over a 24-hour period, aggregated for each hour.
    The black markers in the middle and bottom rows correspond to the time of the top row.
    }\label{fig:inference}
\end{figure}

\begin{figure}[h!]
    \begin{center}
    \includegraphics[width=12cm]{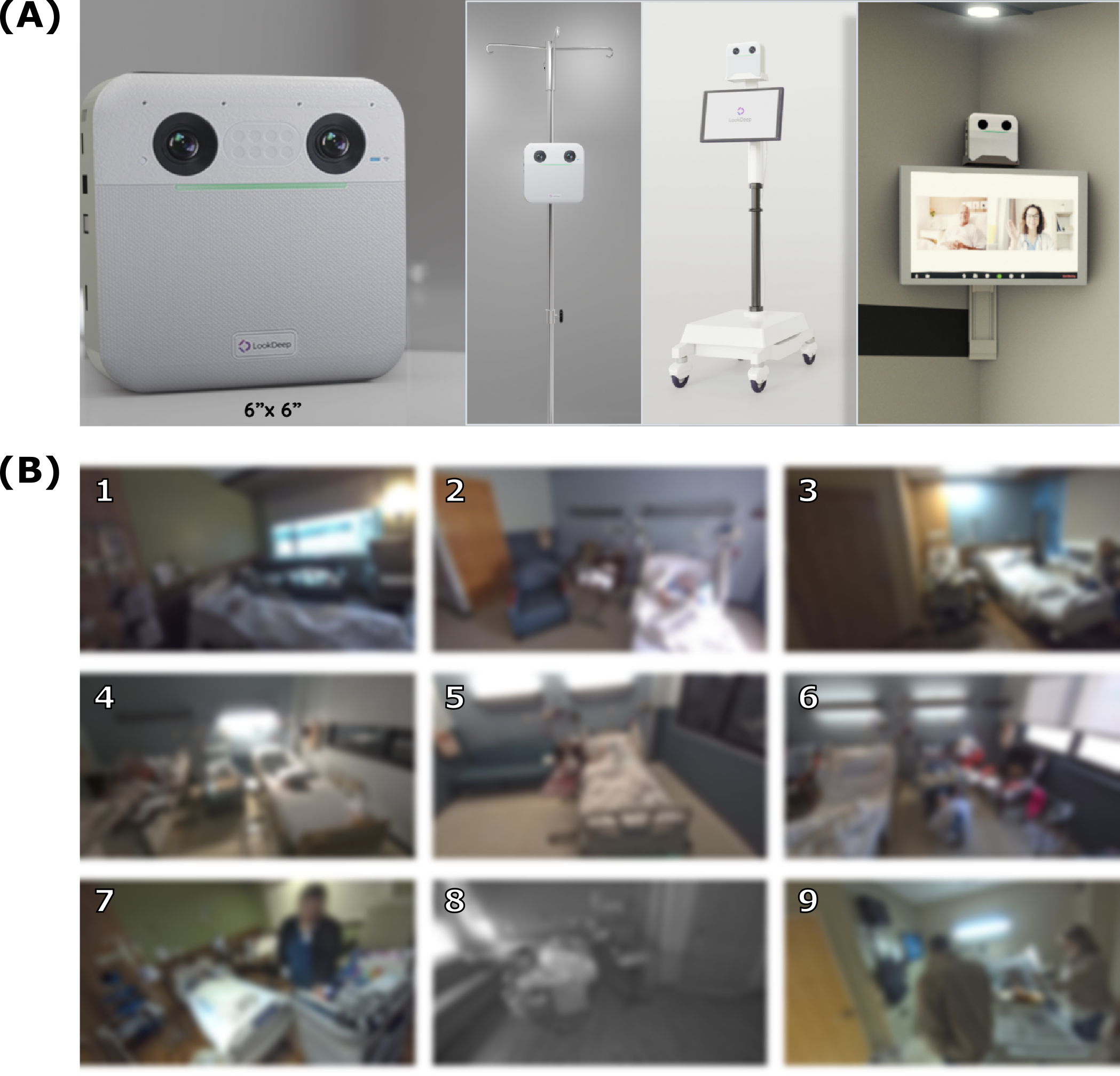}
    \end{center}
    \caption{\textbf{Camera Setup and Example Frames.} 
    (A) LookDeep Video Unit (LVU), a 6” x 6” device, in various mounting configurations. 
    (B) A 3x3 grid of representative frames captured by the system, showing a diversity of configurations. 
    All images are intentionally blurred to maintain privacy.
    Each numbered frame provides a unique example that is found in Figure \ref{fig:camera_meta}.
    }\label{fig:camera}
\end{figure}

The LookDeep Health monitoring system processes video through a computer vision pipeline to detect, classify, and analyze key elements within the patient's room, providing actionable insights to healthcare staff (Figure \ref{fig:inference}). 
Key components include:
\begin{enumerate}
    \item \textbf{Video Data Capture and Preprocessing:} Video data is captured at 1 frame per second (fps) by LookDeep Video Unit (LVU) devices deployed in patient rooms (Figure \ref{fig:camera}A). Data is preprocessed to reduce bandwidth and enable efficient analysis.
    \item \textbf{Object Detection and Localization:} A custom-trained model detects key objects ("person", "bed", "chair") and localizes them with bounding boxes.
    \item \textbf{Person-Role Classification:} Detected "person" objects are further classified as “patient”, “staff”, or “other” by augmenting labels with role-specific information. 
    \item \textbf{Motion Estimation:} Dense optical flow estimates motion between consecutive frames, enabling activity tracking in specific regions (e.g. scene, bed, safety zone).
    \item \textbf{Logical Predictions:} High-level predictions (e.g. “person alone”, “patient supervised by staff”) are derived by applying rules to detection and motion data, with a 5-second smoothing filter to mitigate detection errors.
\end{enumerate}

Inference results, including object detections, role classifications, motion estimation, and logical predictions, are securely stored in a Google cloud database for further analysis (e.g. trend analysis). Anonymized frames are stored at regular intervals for quality assurance and model improvement.

\subsubsection{Data Anonymization}
To ensure patient privacy in accordance with the Health Insurance Portability and Accountability Act (HIPAA) and institutional guidelines, all video data was processed to remove identifiable information. 
For training purposes, frames were face-blurred using a two-step procedure to maintain privacy while preserving relevant scene context:
\begin{enumerate}
    \item \textbf{Manual Labeling:} Faces were manually labeled on fully-blurred images to create bounding boxes without exposing identifiable features. 
    \item \textbf{Local Gaussian Blurring:} A Gaussian blur was applied to labeled facial regions, preserving scene context while anonymizing identities.
\end{enumerate}

This process ensured privacy while enabling effective model training and validation. Data handling was conducted under a Business Associate Agreement (BAA) with participating hospitals. 

\subsection{Data Collection}
\subsubsection{Video Patient Monitoring}
LVU devices capture continuous video in RGB or near-infrared (NIR) mode, depending on ambient lighting. 
Each device is equipped with a CPU and Neural Processing Unit (NPU), capable of processing data at 1fps to minimize latency and reduce cloud processing requirements. 
Inference results are uploaded to a secured cloud database (Google BigQuery), with blurred frames stored separately for manual annotation.
Camera placement varied based on room layout and clinical workflows (Figure \ref{fig:camera}B).

\subsubsection{Annotations}
\begin{figure}[h!]
    \begin{center}
    \includegraphics[width=8cm]{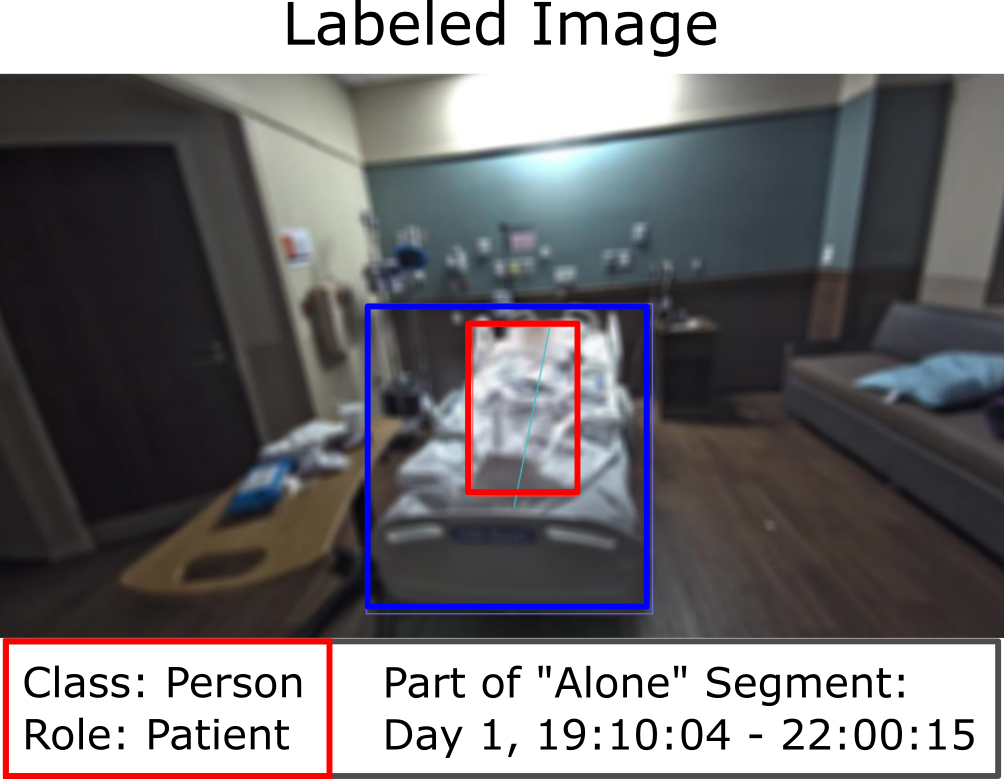}
    \end{center}
    \caption{\textbf{Manually Labeled Image with Observation Log Alignment.} 
    The bed is highlighted with a blue bounding box.
    The patient, identified as a "Person" with the role "Patient", is highlighted with a red bounding box. 
    The associated observation log for "Alone" is shown for illustrative purposes.
    }\label{fig:labels}
\end{figure}

\paragraph{Frame-level Labels}
A professional labeling team manually annotated over 40,000 images with object bounding boxes, object properties, and scene-level tags (Figure \ref{fig:labels}). 
Objects were annotated with 2-d bounding boxes classed as “person”, “bed”, or “chair”, and each "person" bounding box was also assigned a role of “patient”, “staff”, or “other”. 
Scene level attributes were added for whether the patient was “in bed” or “not in bed” and whether the scene included “exception cases” in comparison to stated norms. 
Exception cases were applied in any instance of labeler uncertainty (e.g. difficult to see person, patient in street clothes, etc.); in instances of multiple exception cases being applicable, a single “frame exception” catch-all was used. 
Annotations and quality review were conducted using the Computer Vision Annotation Tool (CVAT, \cite{cvat}), and final QA was conducted using the FiftyOne tool (\cite{fiftyone}).

\paragraph{Observation Logs}
Blurred video summaries for 10 patients (54 patient-days) were reviewed to log periods when the patient was alone. 
Logs included timestamps with 1-2 second precision (Figure \ref{fig:labels}), and underwent secondary quality assurance to provide feedback to labelers and fill out any missing periods. 

\subsection{Computer Vision Predictions}
The LookDeep Health pipeline processes video data using custom-trained models to detect objects, classify person-role, and estimate motion at 1 fps. Preprocessing compresses frames to JPEG at 80\% quality and resizes to a resolution of 1088x612 to reduce bandwidth consumption while still meeting downstream model requirements. Image processing is conducted using OpenCV (\cite{opencv}) and RKNN-toolkit (\cite{rknn}).

\begin{enumerate}
    \item \textbf{Object Detection (Person/Bed/Chair):} Based on the YOLOv4 architecture (\cite{bochkovskiy2020yolov4}), the model identifies key objects in each frame, including “person”, “bed”, and “chair”.
    Training models were initialized using COCO weights (\cite{lin2014microsoft}), then fine-tuned on labeled data. 
    Input images were down-sampled to 608x608 with OpenCV's cubic interpolation method to fit model requirements. 
    Training was conducted on NVIDIA 3070 GPU, and models were subsequently converted for execution on the Rockchip RKNN embedded in the LVU devices.
    \item \textbf{Person Classification (Patient/Staff/Other):} During object detector training, bounding box labels were augmented to classify detected persons by role (“patient”, “staff”, “other”).
    Then, at inference time, each “person-” bounding box are re-labeled as “person”, with the specific role saved in a separate classification field. 
    Confidence scores for role classifications are derived by taking the highest detection confidence as the primary class and distributing residual scores across remaining classes to indicate potential alternate roles.
    \item \textbf{Optical Flow (Motion Estimation):} Motion between frames was estimated using the Gunnar-Farneback dense optical flow algorithm, which calculates horizontal and vertical displacement for each pixel (\cite{farneback2003two}). 
    Optical flow inputs were converted to grayscale and down-sampled to 480x270 to ensure real-time execution. 
    For each region of interest, average motion magnitude was calculated by averaging horizontal and vertical flow vectors, providing an indicator of activity intensity. 
    This estimation does not require training and was implemented using OpenCV with fixed parameters: pyramid scale (pyr\_scale=0.5), number of pyramid levels (levels=3), window size (winsize=15), number of iterations (iterations=3), size of pixel neighborhood used to find polynomial expansion (poly\_n=5), and the standard deviation of the Gaussian used to smooth derivatives (poly\_sigma=1.2).
\end{enumerate}

\subsubsection{Additional Components}
\paragraph{Regions of Interest (ROIs)}
ROIs, such as "safety zones", provide contextual boundaries for monitoring. 
They are not predictive outputs themselves, but instead are used to track patient movements and boundary crossings.
The "safety zone" was a polygonal region defined by the virtual monitor; its pixel mask is generated by expanding the boundary perimeter by 10\% to ensure effective monitoring.
Additional ROIs used by the system include the full scene and the detected bed. 

\paragraph{Logical Predictions}
Logical predictions summarize patient status and interactions. 
These predictions were derived from a combination of object detection and role classification results and smoothed with a 5-second filter to mitigate intermittent detection errors.
\begin{itemize}
    \item \textbf{Person Alone:} \textit{True} when the average number of detected people in the room is less than two.
    \item \textbf{Patient Alone:} \textit{True} when the average number of detected people in the room is less than two, and at least one person is classified as a patient.
    \item \textbf{Supervised by Staff:} \textit{True} when the average number of detected people in the room is two or more, and at least one person is classified as healthcare staff.
\end{itemize}

\paragraph{Trend Predictions}
Trends provide insights into immediate and long-term patient activity, aiding in risk identification and care planning. 
Hourly trends summarize patient behavior (e.g. "alone" or "moving") based on aggregated logical predictions. 
For each one-hour interval, predictions were used to calculate the percentage of time the patient spent in key states like "alone," "supervised by staff," or "moving".
These percentages were then plotted over time to visualize hourly trends in patient isolation or activity levels throughout the day.
These trends provide a high-level overview of patient behavior, aiding in the identification of potential risks and informing care decisions.

\subparagraph{"Assisted" Trend Predictions}
A one-off analysis was conducted to simulate the system's performance when one of the predictions was known. The system's trend predictions based solely on AI inference were compared with those generated using a combination of AI inference and observation logs. 
For this comparison, "assisted" trends were created by integrating AI-predicted states for "moving" and "supervised by staff" with manually logged periods of "alone" status from the observation logs. 

\subsection{Evaluation}
The performance of the AI-driven monitoring system was assessed through two primary methods: \textbf{image-level assessment} and \textbf{comparison against observation logs}. 
In the image-level assessment, each frame was analyzed against manual annotations to evaluate the accuracy of the system's object detection, person-role classification, and scene interpretation capabilities. 
In parallel, observation logs, created from anonymized video summaries of select patients, were compared against predicted trends to assess the system's ability to capture patient behavior patterns.

\subsubsection{Frame-level Analysis}
Each model in the AI system was evaluated independently to assess its performance in object detection and classification tasks. 
Key performance metrics—precision, recall, and F1-score—were calculated to measure the accuracy and reliability of each model's predictions. 
Precision assessed the proportion of true positives among all predicted positives, recall measured the ability to identify all true positives, and the F1-score provided a balanced metric between precision and recall. 

In addition to these direct object detection and classification tasks, the AI system also generated higher-level, “logical” predictions derived from these outputs.
For example, the prediction “is patient alone” was inferred based on a combination of object detection results, such as the absence of healthcare staff within a defined proximity to the patient. 
These logical predictions were treated as classification tasks themselves, with their accuracy similarly evaluated using precision, recall, and F1-score metrics based on labeled image data. 
This multi-layered approach allowed us to thoroughly validate both the core object detection functions of each model and the system's ability to interpret and apply these outputs to patient monitoring tasks.

\subsubsection{Trend Analysis}
Trend analysis was conducted by comparing the system's inference-derived metrics to ground truth metrics recorded in observation logs, with both datasets aggregated by patient-day. 
Unlike the hourly trends shown in Figure \ref{fig:inference}, analysis was conducted at the per-second level to ensure accurate alignment between AI predictions and observation logs.
The primary metric for this analysis was logistic regression accuracy, which assessed the AI system's ability to predict observed behaviors within three time periods: daytime (6 am to 9 pm), nighttime (9 pm to 6 am), and the full 24-hour period. 
In cases where only a single class (e.g. "alone" or "not alone") was present within a specific time period, logistic regression was not feasible. 
Instead, a manual accuracy score was computed, to allow for consistent accuracy measurements across all time intervals.
This score is defined as the proportion of matching values between the AI predictions and ground truth. 

Focusing on the "alone" binary behavior trend enables an assessment of the alignment between AI predictions and real-world observations. 
This analysis validated the AI system's effectiveness in capturing hourly patient behavior trends, underscoring its potential utility in real-time patient monitoring and early detection of deviations from expected patterns.

\subsubsection{Camera Position Meta-analysis}
\begin{figure}[h!]
    \begin{center}
    \includegraphics[width=14cm]{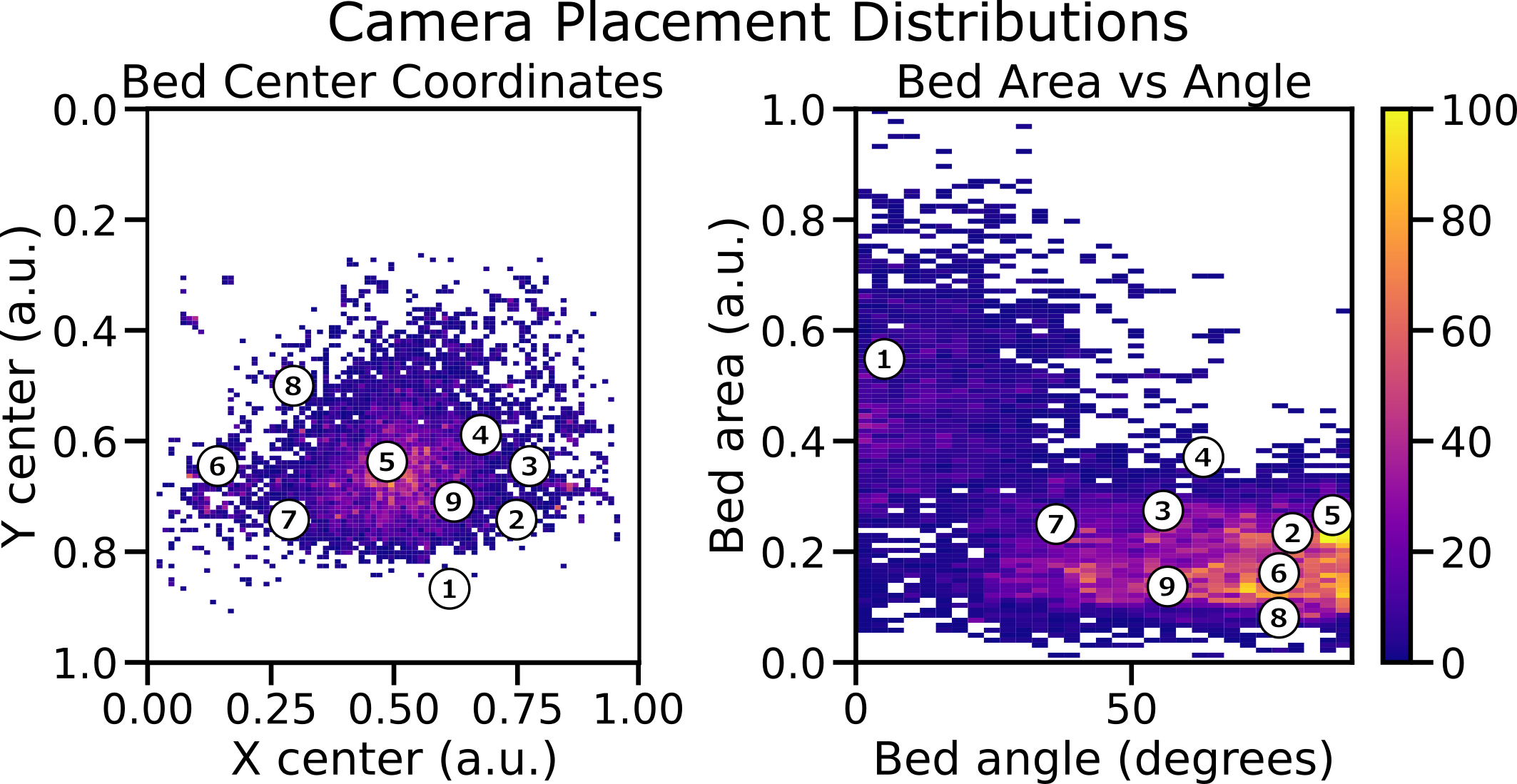}
    \end{center}
    \caption{\textbf{Distribution of Labeled Bed Positions Relative to the Camera.} 
    Left: Spatial variability of bed center coordinates. 
    Right: Distribution of bed area versus bed angle relative to the camera.
    Each numbered point is shown in Figure \ref{fig:camera}B.
    This highlights the variations in camera perspective and placement across the study. 
    }\label{fig:camera_meta}
\end{figure}

Since cameras were mounted on mobile carts rather than fixed positions, there was variability in camera setup across patients and hospital rooms (Figure \ref{fig:camera}B). 
To explore the potential impact of this variability, labeled bed locations were used to estimate each camera's position relative to the hospital bed. 
Distributions of the labeled bed area and size within each frame, along with the centroid location of the bed relative to the camera's field of view are plotted in Figure \ref{fig:camera_meta}. 
These distributions provide an indirect measure of camera position.

This exploratory analysis helped identify patterns and variations in camera setups across different monitoring sessions. 
However, this information was observational and used only to understand positional variability; no specific adjustments were made during model training or evaluation to account for different camera positions. 
The results underscore the robustness of our models in handling diverse camera perspectives, as the system maintained consistent detection performance despite these variations.

\section{Results}
\subsection{Frame-level Analysis}
\subsubsection{Object Detection, Role Identification, and Patient Isolation Classification}
The evaluations demonstrated that the custom-trained computer vision models perform robustly in real-world hospital settings, achieving high precision across key object detection and classification tasks. 
We compared five production models alongside a baseline model using an off-the-shelf YOLOv4 configuration (Table \ref{table:object_detection}). 
Each production model corresponds to a different release, with progressively larger and more refined training datasets incorporated over time (Figure \ref{fig:timeline}). 
This iterative refinement led to increased model accuracy and adaptability in real-world hospital settings.
To ensure consistency, all frame-level analysis was conducted on 10,000 frames collected over a two year period.
This representative sample, excluded from model training and validation, highlights the incremental improvements achieved by expanding training datasets across model versions.

As newer models were released, the training set was expanded to include additional annotated data, allowing each successive model to capture more complex and diverse scenarios encountered in hospital environments. 
The most recent fine-tuned model (v5) achieved an \textbf{F1-score of 0.91} for detecting “person”, notably surpassing the baseline YOLOv4 model score of 0.41 (Table \ref{table:remaining_metrics}). 
Across all object classes—including beds, furniture, and other room elements—the v5 model demonstrated an \textbf{F1-score of 0.92}, reflecting a high degree of accuracy and consistency across diverse object types.

In addition to object detection, the system was evaluated on a three-class person-role classification task, distinguishing between patients, healthcare staff, and visitors within the camera's field of view. 
The v5 model demonstrated particularly strong performance for the "patient" class, achieving an \textbf{F1-score of 0.98}, which reflects its high accuracy in identifying patients specifically (Table \ref{table:remaining_metrics}). 
Accurate person-role classification is essential for monitoring patient interactions and ensuring appropriate caregiving behaviors, as it enables the system to capture not only the presence of individuals but also their roles. 
Focusing on the "patient" class, the high F1-score underscores the model's robustness in tracking patient activity and interactions, which are critical for effective continuous monitoring in dynamic hospital environments.

The downstream classification task of identifying whether a patient was “alone” in the room showed similarly strong results, with the v5 model achieving an \textbf{F1-score of 0.92} (Table \ref{table:remaining_metrics}). 
This classification task, essential for monitoring patient isolation, consistently improved with each new production release, as more comprehensive training data contributed to better model accuracy. 
These results confirm the advantage of iterative model refinement and dataset expansion, with each production release yielding models that are better adapted to the variability and demands of real-world hospital settings.

\subsubsection{Impact of Privacy-Preserving Blurring on Model Consistency}
The performance consistency of the models across unblurred and face-blurred images was evaluated using the $\Delta$ metric, which represents the F1-score difference between the two image types (Table \ref{tab:blurred_vs_unblurred}). 
Across all model versions, the $\Delta$ values were relatively small, indicating that face-blurring—a common privacy-preserving preprocessing step—had minimal impact on model accuracy. 
For versions v3 and v4, the $\Delta$ value was +0.04, while in v5 it decreased to +0.02, suggesting improved robustness to blurring as the training data volume increased.

A smaller $\Delta$ value is desirable as it indicates that the model performs consistently regardless of whether the images are unblurred or face-blurred. 
The reduction in $\Delta$ for v5 highlights the value of larger, more diverse training datasets in ensuring that the models generalize well across different image types. 
This is particularly important in hospital settings, where preserving patient privacy often necessitates the use of face-blurred images. 
The ability to maintain high accuracy in such scenarios ensures the system's practicality and reliability for real-world deployment.

These results demonstrate that the models not only achieve high accuracy but also exhibit resilience to variations introduced by privacy-preserving preprocessing, a key requirement for scalable applications in healthcare environments.

\subsection{Trend Analysis}
\begin{figure}[h!]
    \begin{center}
    \includegraphics[width=8cm]{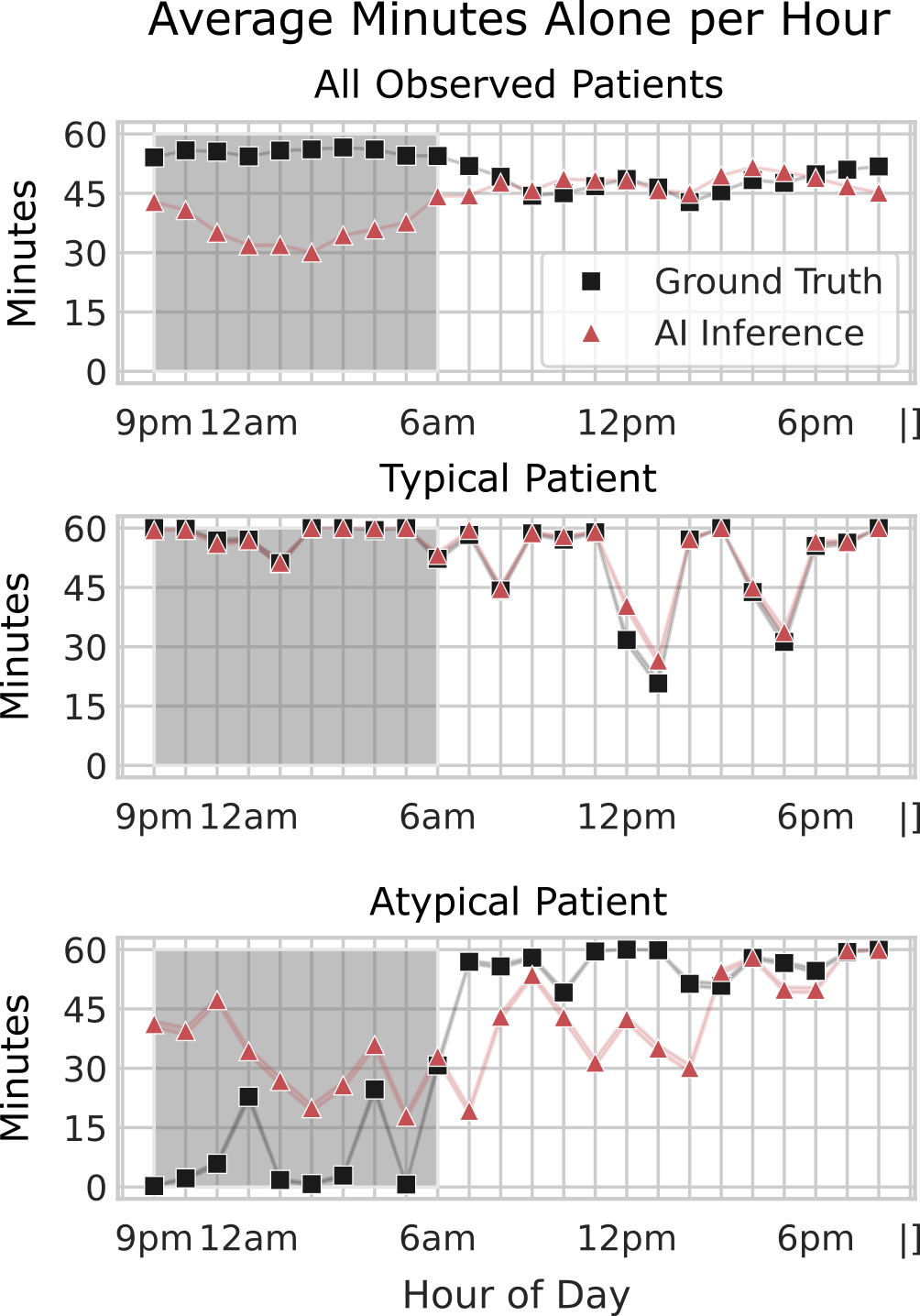}
    \end{center}
    \caption{\textbf{Comparison of Average Minutes Alone per Hour Across All Patients, a Typical Patient, and an Atypical Patient.} 
    The average minutes patients spent alone per hour, comparing ground truth (black squares) and AI inference (red triangles) across three scenarios: all patients (top), a typical patient (middle), and an atypical patient (bottom). 
    The x-axis shows the hour of the day, while the y-axis indicates the average minutes alone per hour. 
    The shaded region represents nighttime hours (9 pm to 6 am). 
    For all patients, AI inference closely aligns with ground truth during the day but shows less accuracy at night. 
    An example of a typical and an atypical patient is shown to illustrate the variability in alone time patterns across individual patients.
    Unlike average trends, this atypical patient exhibits an overprediction of alone time at night, highlighting the need for further model refinement to capture individual patient behaviors accurately.
    }\label{fig:eval_obs}
\end{figure}

Inference-derived trends for the “patient alone” metric were compared against observation logs to evaluate the system's ability to accurately capture real-world patterns (Figure \ref{fig:eval_obs}). 
This trend analysis utilized data from earlier stages of the project when base models with lower performance were deployed. 
Specifically, the object detectors used for these inferences had an F1-score of 0.85 for “person” detection, which is lower than the performance of the latest models. 
Despite this, the analysis showed strong alignment with ground truth data, achieving an \textbf{average logistic regression/manual accuracy of 0.84 ± 0.13 during daytime, 0.80 ± 0.16 at nighttime, and 0.82 ± 0.15 across all times}. 
These results highlight the robustness of the AI system in capturing patient isolation trends, even when using earlier model versions with lower baseline performance.

\begin{figure}[h!]
    \begin{center}
    \includegraphics[width=8cm]{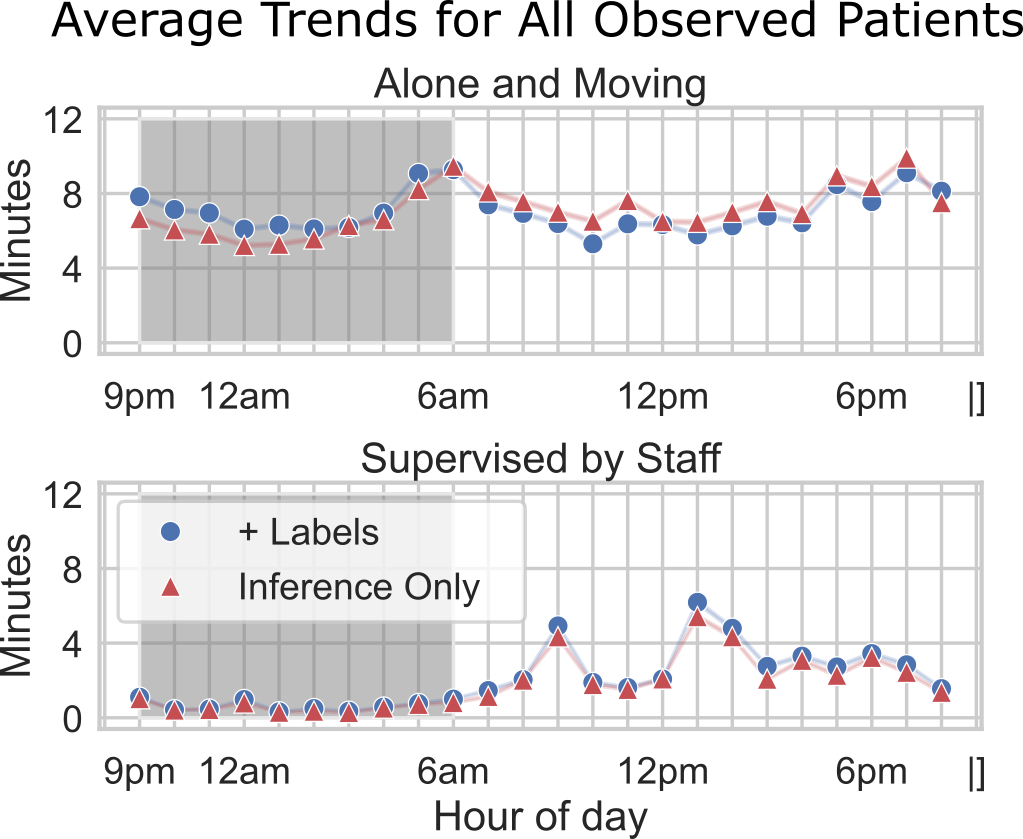}
    \end{center}
    \caption{\textbf{Average Trends for All Observed Patients.} 
    Hourly trends are compared across two metrics: Alone and Moving (top) and Supervised by Staff (bottom).
    Fully AI-inferred data ("Inference Only") is plotted with red triangles.
    "Assisted" data ("+Labels") is plotted with blue circles - for these data, ground truth "alone" status was used instead of AI-inference. 
    The x-axis indicates the hour of the day, while the y-axis shows the average minutes per hour. 
    The shaded region represents nighttime hours (9 pm to 6 am). 
    Although there is a discrepancy between labeled and inferred data for Alone, particularly during nighttime hours, the downstream impact on overall trend accuracy appears minimal (1-2 minutes per hour). 
    }\label{fig:eval_trends}
\end{figure}

This accuracy indicates that, even with slightly reduced detection precision in the older models, the system could reliably capture general patterns in patient isolation behavior. 
The standard deviation (± 0.15) reflects some variability in accuracy across different times of day and patient conditions, possibly influenced by factors such as changing camera angles or environmental conditions. 
As shown in the normative hourly trends (Figure \ref{fig:eval_trends}), discrepancies between labeled and AI-inferred "alone" data are more pronounced during nighttime hours, but these differences have minimal impact on the broader trend patterns. 
For both "Alone and Moving" and "Supervised by Staff" metrics, the AI inferences closely align with label-assisted data, amounting to an \textbf{average error of 1-2 minutes per hour}. 
This consistency underscores the model's robustness in capturing meaningful patient-alone trends and suggests that any nighttime performance gaps in the "alone" inference do not significantly compromise the overall accuracy. 
These results highlight the model's potential for improved trend detection as newer, refined models are applied to subsequent data.

\section{Discussion}
\subsection{Implications for Clinical Practice}
The findings of this study underscore the potential for AI-enabled patient monitoring systems to enhance clinical practice through continuous, real-time monitoring. 
Traditional in-person observations are limited by the time constraints of healthcare staff, who spend limited hours directly interacting with each patient. 
By providing continuous monitoring, the LookDeep Health platform enables staff to detect patterns that would otherwise go unnoticed, such as extended periods of patient isolation, movement patterns that might indicate a risk of falls, pressure injuries, or irregular interactions with staff. 
Real-time alerts based on these observations could prompt timely interventions, potentially improving patient safety and outcomes.

Moreover, the data collected by this system can inform trend analysis on a population level, supporting hospital resource allocation and staffing decisions. 
For instance, identifying times of day when patients are frequently unsupervised could guide adjustments in staffing or the deployment of additional monitoring resources to high-risk patients. 
Beyond staffing, the system's insights into patient mobility patterns—such as time spent in bed, in a chair, or walking around the room—can help identify markers of successful recovery and readiness for discharge, contributing to improved patient outcomes. 
These mobility insights could also support the development of best practices for post-procedure mobility, tailored to specific surgeries or treatments, to enhance patient recovery. 
Altogether, these data-driven insights promote a more efficient, personalized approach to patient care, potentially improving patient satisfaction and clinical outcomes.

\subsection{Impact of Face-Blurring on Model Performance}
While the evaluation of model performance on both unblurred and face-blurred images provides valuable insights, it is important to note that face-blurring is applied only during training and evaluation phases. 
In real-world deployment, the model will encounter unblurred images as it monitors patients in hospital settings, making this distinction critical to understanding its practical performance. 
The small $\Delta$ values observed across different model versions indicate that the models have been designed to handle face-blurred images without significant degradation in performance. 
The reduced $\Delta$ in the latest version (v5), attributed to increased training data volume, demonstrates improved resilience to face-blurring.
However, further studies are needed to assess the model's performance in unblurred scenarios, particularly in environments where face-blurring images for training and evaluation is not an option.
This approach ensures privacy during development while maintaining practical deployment fidelity, as real-time monitoring operates on unblurred frames.

\subsection{Variation in Camera Setup}
The LookDeep Health patient monitoring platform was deployed in real-world hospital settings with cameras mounted on mobile carts rather than fixed positions, resulting in variation in camera angles, distances, and perspectives across different patient rooms. 
This variability introduced potential challenges in maintaining consistent object detection and classification accuracy, as model performance can be influenced by changes in camera field of view and positioning relative to the bed. 
To mitigate these effects, we conducted a camera position meta-analysis using metadata on labeled bed area and centroid location to estimate the approximate camera placement within each room. 
Our analysis confirmed that, despite positional differences, the model consistently achieved reliable performance across object detection and classification tasks, demonstrating its robustness to spatial variability. 
However, this setup presents limitations in controlling for optimal camera positioning, a factor that future studies with standardized camera setups could explore further to minimize variability and enhance model reliability.

\subsection{Nuanced Differences in Time Coverage of Analyses}
A key aspect of this study is the variation in time coverage across different datasets, reflecting the evolving nature of data collection and model validation in real-world hospital settings. 
The observation logs dataset, which provided ground truth for logical trend validation, was collected exclusively in 2023. 
In contrast, frame-level annotations for evaluating object detection and person-role classification were gathered over a more extended period from 2022 to 2024. 
Additionally, the publicly released dataset comprises data collected from a 6 month span across 2024, representing over 1,000 collective patient days across multiple hospitals.

These differences in collection periods introduce nuances in interpretation. 
For instance, frame-level evaluations benefit from the broader time span, capturing a variety of hospital conditions and patient behaviors across seasons and changing workflows. 
However, trend analyses were constrained to the observation log time frame, which may limit the ability to generalize trends across the entire study period. 
Similarly, the released dataset reflects data from the latter phase of the study, aligning with the most refined models but excluding early-stage model iterations.

These variations in time coverage highlight the need to contextualize each analysis within its specific time frame. 
Future studies could benefit from aligning data collection periods across all evaluation methods, ensuring that models validated on frame-level tasks are continuously validated against trend and behavioral analyses for consistent performance insights over time.

\subsection{Challenges and Limitations}
Several challenges and limitations were encountered in this study. 
First, the variability in camera setup, as mentioned earlier, introduces potential inconsistencies in model performance due to changing perspectives and distances. 
While our metadata analysis mitigated this to some extent, a standardized camera setup would likely yield more consistent results.

Second, while the LookDeep Health system demonstrated strong performance in object detection and role classification, real-time video processing presents computational challenges that require balancing accuracy and processing speed. 
Our use of onboard CPU and NPU on LVU devices provided sufficient processing capabilities for 1 fps inference; however, the scalability of such a setup may be constrained in larger hospital systems requiring higher frame rates for finer details.

Third, the dataset collected in this study primarily consists of high-risk fall patients, which may limit the generalizability of findings to broader patient populations - for example, high-risk patients exhibit limited mobility compared to other patient groups.
Additionally, the analysis was conducted on older model versions for some trend analyses, potentially lowering the accuracy of trend detection. 
Although model refinements are expected to improve results, these differences in model versions should be considered when interpreting the findings.

Lastly, maintaining patient privacy is paramount in continuous video monitoring systems. 
While the LookDeep Health platform anonymizes all video and stores de-identified data, ongoing attention to data privacy and compliance with healthcare regulations is essential for future deployments in clinical environments.

\subsection{Suggestions for Future Research}
While this study provides a foundation for understanding the impact of AI-driven patient monitoring, further research is warranted to explore additional facets of this technology. 
Future studies could investigate:
\begin{itemize}
    \item \textbf{Advanced Deep Learning Techniques:} Integrating more sophisticated deep learning architectures could enhance the detection of subtle anomalies, while adaptive pipelines could improve real-time robustness in dynamic hospital environments.
    \item \textbf{Refining Architectures and Guardrails:} Future work could involve refining architectures to detect edge cases more accurately, tracking patterns in prediction errors, and incorporating confidence-based guardrails to prevent catastrophic failures. 
    Such guardrails could include alerts when model confidence is unexpectedly low for consecutive predictions.
    \item \textbf{Higher Frame Rates and Computational Scaling:} Evaluating the potential for higher frame rates or adaptive frame rate technology to improve real-time responsiveness, particularly in high-activity environments.
    \item \textbf{Standardization of Camera Placement:} Testing standardized, fixed camera setups across patient rooms aims to minimize positional variability and improve model consistency. 
    Although standardization can reduce variability, embracing the inherent diversity of setups may enhance model robustness for real-world applications.
    \item \textbf{Expanded Patient Cohorts:} Extending the analysis to include a wider range of patient demographics and conditions to assess generalizability and adapt the system to diverse populations.
    \item \textbf{Interoperability with Hospital Systems:} Integrating AI-driven monitoring with electronic health records (EHRs) and hospital workflow systems to provide context-aware alerts and streamline clinical response.
\end{itemize}

These research directions, alongside continued refinement of computer vision models and monitoring systems, will be essential for advancing the practical application of AI in patient monitoring and driving further improvements in healthcare delivery.

\section{Conclusion}
AI integration in medical imaging is advancing personalized patient treatment but still faces challenges related to effectiveness and scalability. 
This work demonstrates the potential of computer vision as a foundational technology for continuous and passive patient monitoring in real-world hospital environments.

The contributions of this study are two-fold. 
First, we introduce the LookDeep Health patient monitoring platform, which leverages computer vision models to monitor patients continuously throughout their hospital stay. 
This platform scales to support a large number of patients and is designed to handle the complexities of hospital-based data collection. 
Using this system, we have compiled a unique dataset of computer vision predictions from over 300 high-risk fall patients, spanning 1,000 collective days of monitoring. 
To encourage further exploration in the field, we released this anonymized dataset publicly at \href{https://github.com/lookdeep/ai-norms-2024}{lookdeep/ai-norms-2024}.

Second, we rigorously validated the AI system, demonstrating strong performance in image-level object detection and person-role classification tasks. 
Our analysis also confirms a positive alignment between inference-derived trends and human-observed behaviors on a patient-hour basis, underscoring the reliability of the AI system in capturing patient activity trends. This evaluation can serve as a benchmark for future studies, providing a standard set of criteria for assessing the performance of AI-driven patient monitoring systems.

The extensive dataset and rigorous validation of the LookDeep Health platform highlight the feasibility and impact of continuous patient monitoring through video. 
By offering real-time insights into patient activity and isolation patterns, continuous monitoring has the potential to reduce fall risks by alerting staff to high-risk situations as they unfold. 
Beyond improving patient safety, these insights support more efficient staffing and resource allocation, allowing hospitals to adjust care based on real-time patient needs. 
This predictive capability also aids administrators in managing bed occupancy and optimizing patient flow, particularly during peak times, thus enhancing the responsiveness, efficiency, and scalability of the healthcare system. 
This work paves the way for future advancements in AI-driven healthcare solutions, promising scalable, data-informed insights to elevate patient care and hospital management.

\section*{Conflict of Interest Statement}
The authors disclose the following competing financial interests: they are current or former employees of LookDeep Health, the company that provided the tools used in this study. 
LookDeep Health was involved in data collection and analysis and reviewed the final manuscript prior to submission. 
The authors declare no other conflicts of interest related to this work.

\section*{Research Ethics}
The data used in this retrospective study was collected from patients admitted to one of eleven hospital partners across three different states in the USA.
The study and handling of data followed the guidelines provided by CHAI standards.
Access to this data was granted to the researchers through a Business Associate Agreement (BAA) specifically for monitoring patients at high risk of falls.
In compliance with the Health Insurance Portability and Accountability Act (HIPAA), patients provided written informed consent for monitoring as part of their standard inpatient care. 
To ensure patient privacy, all video data was blurred prior to storage, and no identifiable information is included in this work. 
Face-blurred frames were used only for training purposes. 
Faces were manually labeled on fully-blurred images, and the raw images were then treated with a local Gaussian blur in the facial regions, ensuring privacy without compromising model training quality. 
The outcomes of this analysis did not influence patient care or clinical outcomes. 

\section*{Author Contributions}
P.G. drafted the manuscript. 
P.G., T.T., and N.S. designed the research. 
P.G., P.R., and T.T. conducted the research and analyzed the data. 
T.W. and M.C. provided clinical insights. 
N.S. supervised the study.

\section*{Funding}
This research was funded by our hospital system partner as part of a business agreement supporting the development and deployment of AI-driven patient monitoring solutions. 
The funding provided resources for data collection, system implementation, and analysis within the hospital environment.

\section*{Acknowledgments}
First and foremost, we extend our gratitude to additional members of the LookDeep Health team, both past and present—Guram Kajaia, James Eitzman, Bill Mers, Mike O'Brien, Jan Marti, Laura Urbisci, and Tom Hata—for their work in building the patient monitoring platform. 
We acknowledge the assistance of OpenAI's ChatGPT (version 4, model GPT-4-turbo) in refining the text of this manuscript. 
This generative AI technology was accessed through OpenAI's platform and used to improve clarity and organization in the presentation of the research.
Finally, we thank A.P., K.C., and T.P. for their valuable feedback on the manuscript.
Please note this version of the manuscript is a pre-print submitted to arXiv. 

\section*{Data Availability Statement}
To support reproducibility and encourage further exploration in the field, we released a publicly available dataset along with the associated code at \href{https://github.com/lookdeep/ai-norms-2024}{lookdeep/ai-norms-2024}. 
This dataset consists of anonymized video-based computer vision predictions from over 300 high-risk fall patients monitored over 1,000 collective days in real-world hospital environments. 
The accompanying code includes scripts for data processing and analysis, enabling others to replicate our methods and validate our findings.
We invite researchers and practitioners to utilize this resource for advancing continuous patient monitoring technologies and exploring new applications in AI-driven healthcare.

\bibliographystyle{Frontiers-Harvard}
\bibliography{references}

\begin{thebibliography}{24}
\providecommand{\natexlab}[1]{#1}
\expandafter\ifx\csname urlstyle\endcsname\relax
  \providecommand{\doi}[1]{doi:\discretionary{}{}{}#1}\else
  \providecommand{\doi}{doi:\discretionary{}{}{}\begingroup \urlstyle{rm}\Url}\fi
\providecommand{\selectlanguage}[1]{\relax}
\providecommand{\bibAnnoteFile}[1]{%
  \IfFileExists{#1}{\begin{quotation}\noindent\textsc{Key:} #1\\
  \textsc{Annotation:}\ \input{#1}\end{quotation}}{}}
\providecommand{\bibAnnote}[2]{%
  \begin{quotation}\noindent\textsc{Key:} #1\\
  \textsc{Annotation:}\ #2\end{quotation}}

\bibitem[{Abbe and O'Keeffe(2021)}]{abbe2021continuous}
Abbe, J.~R. and O'Keeffe, C. (2021).
\newblock Continuous video monitoring: Implementation strategies for safe patient care and identified best practices.
\newblock \emph{Journal of nursing care quality} 36, 137--142
\bibAnnoteFile{abbe2021continuous}

\bibitem[{Avogaro et~al.(2023)Avogaro, Cunico, Rosenhahn, and Setti}]{avogaro2023markerless}
Avogaro, A., Cunico, F., Rosenhahn, B., and Setti, F. (2023).
\newblock Markerless human pose estimation for biomedical applications: a survey.
\newblock \emph{Frontiers in Computer Science} 5, 1153160
\bibAnnoteFile{avogaro2023markerless}

\bibitem[{Bajwa et~al.(2021)Bajwa, Munir, Nori, and Williams}]{bajwa2021artificial}
Bajwa, J., Munir, U., Nori, A., and Williams, B. (2021).
\newblock Artificial intelligence in healthcare: transforming the practice of medicine.
\newblock \emph{Future healthcare journal} 8, e188--e194
\bibAnnoteFile{bajwa2021artificial}

\bibitem[{Bochkovskiy et~al.(2020)Bochkovskiy, Wang, and Liao}]{bochkovskiy2020yolov4}
[Dataset] Bochkovskiy, A., Wang, C.-Y., and Liao, H.-Y.~M. (2020).
\newblock Yolov4: Optimal speed and accuracy of object detection
\bibAnnoteFile{bochkovskiy2020yolov4}

\bibitem[{Bradski(2000)}]{opencv}
Bradski, G. (2000).
\newblock {The OpenCV Library}.
\newblock \emph{Dr. Dobb's Journal of Software Tools}
\bibAnnoteFile{opencv}

\bibitem[{Capo-Lugo et~al.(2023)Capo-Lugo, Young, Farley, Aquino, McLaughlin, Colantuoni et~al.}]{capo2023revealing}
Capo-Lugo, C.~E., Young, D.~L., Farley, H., Aquino, C., McLaughlin, K., Colantuoni, E., et~al. (2023).
\newblock Revealing the tension: The relationship between high fall risk categorization and low patient mobility.
\newblock \emph{Journal of the American Geriatrics Society} 71, 1536--1546
\bibAnnoteFile{capo2023revealing}

\bibitem[{Chae et~al.(2021)Chae, Choi, Park, and Jang}]{chae2021improved}
Chae, W., Choi, D.-W., Park, E.-C., and Jang, S.-I. (2021).
\newblock Improved inpatient care through greater patient--doctor contact under the hospitalist management approach: A real-time assessment.
\newblock \emph{International journal of environmental research and public health} 18, 5718
\bibAnnoteFile{chae2021improved}

\bibitem[{Chen et~al.(2018)Chen, Gabriel, Alasfour, Gong, Doyle, Devinsky et~al.}]{chen2018patient}
Chen, K., Gabriel, P., Alasfour, A., Gong, C., Doyle, W.~K., Devinsky, O., et~al. (2018).
\newblock Patient-specific pose estimation in clinical environments.
\newblock \emph{IEEE journal of translational engineering in health and medicine} 6, 1--11
\bibAnnoteFile{chen2018patient}

\bibitem[{Corporation(2023)}]{cvat}
[Dataset] Corporation, C. (2023).
\newblock Computer vision annotation tool (cvat).
\newblock \doi{10.5281/zenodo.8416684}
\bibAnnoteFile{cvat}

\bibitem[{Davenport and Kalakota(2019)}]{davenport2019potential}
Davenport, T. and Kalakota, R. (2019).
\newblock The potential for artificial intelligence in healthcare.
\newblock \emph{Future healthcare journal} 6, 94--98
\bibAnnoteFile{davenport2019potential}

\bibitem[{Davoudi et~al.(2019)Davoudi, Malhotra, Shickel, Siegel, Williams, Ruppert et~al.}]{davoudi2019intelligent}
Davoudi, A., Malhotra, K.~R., Shickel, B., Siegel, S., Williams, S., Ruppert, M., et~al. (2019).
\newblock Intelligent icu for autonomous patient monitoring using pervasive sensing and deep learning.
\newblock \emph{Scientific reports} 9, 8020
\bibAnnoteFile{davoudi2019intelligent}

\bibitem[{Esteva et~al.(2021)Esteva, Chou, Yeung, Naik, Madani, Mottaghi et~al.}]{esteva2021deep}
Esteva, A., Chou, K., Yeung, S., Naik, N., Madani, A., Mottaghi, A., et~al. (2021).
\newblock Deep learning-enabled medical computer vision.
\newblock \emph{NPJ digital medicine} 4, 5
\bibAnnoteFile{esteva2021deep}

\bibitem[{Farneb{\"a}ck(2003)}]{farneback2003two}
Farneb{\"a}ck, G. (2003).
\newblock Two-frame motion estimation based on polynomial expansion.
\newblock In \emph{Image Analysis: 13th Scandinavian Conference, SCIA 2003 Halmstad, Sweden, June 29--July 2, 2003 Proceedings 13} (Springer), 363--370
\bibAnnoteFile{farneback2003two}

\bibitem[{Fuzhou Rockchip Electronics~Co.(2024)}]{rknn}
[Dataset] Fuzhou Rockchip Electronics~Co., L. (2024).
\newblock Rknn-toolkit.
\newblock Available from https://github.com/rockchip-linux/rknn-toolkit
\bibAnnoteFile{rknn}

\bibitem[{Haque et~al.(2020)Haque, Milstein, and Fei-Fei}]{haque2020illuminating}
Haque, A., Milstein, A., and Fei-Fei, L. (2020).
\newblock Illuminating the dark spaces of healthcare with ambient intelligence.
\newblock \emph{Nature} 585, 193--202
\bibAnnoteFile{haque2020illuminating}

\bibitem[{Lin et~al.(2014)Lin, Maire, Belongie, Hays, Perona, Ramanan et~al.}]{lin2014microsoft}
Lin, T.-Y., Maire, M., Belongie, S., Hays, J., Perona, P., Ramanan, D., et~al. (2014).
\newblock Microsoft coco: Common objects in context.
\newblock In \emph{Computer Vision--ECCV 2014: 13th European Conference, Zurich, Switzerland, September 6-12, 2014, Proceedings, Part V 13} (Springer), 740--755
\bibAnnoteFile{lin2014microsoft}

\bibitem[{Lindroth et~al.(2024)Lindroth, Nalaie, Raghu, Ayala, Busch, Bhattacharyya et~al.}]{lindroth2024applied}
Lindroth, H., Nalaie, K., Raghu, R., Ayala, I.~N., Busch, C., Bhattacharyya, A., et~al. (2024).
\newblock Applied artificial intelligence in healthcare: A review of computer vision technology application in hospital settings.
\newblock \emph{Journal of Imaging} 10, 81
\bibAnnoteFile{lindroth2024applied}

\bibitem[{Mascagni et~al.(2022)Mascagni, Alapatt, Sestini, Altieri, Madani, Watanabe et~al.}]{mascagni2022computer}
Mascagni, P., Alapatt, D., Sestini, L., Altieri, M.~S., Madani, A., Watanabe, Y., et~al. (2022).
\newblock Computer vision in surgery: from potential to clinical value.
\newblock \emph{npj Digital Medicine} 5, 163
\bibAnnoteFile{mascagni2022computer}

\bibitem[{Moore and Corso(2024)}]{fiftyone}
[Dataset] Moore, B.~E. and Corso, J.~J. (2024).
\newblock Fiftyone.
\newblock Available from https://www.voxel51.com/fiftyone/ and https://github.com/voxel51/fiftyone
\bibAnnoteFile{fiftyone}

\bibitem[{Parker et~al.(2022)Parker, Gilstrap, Bedoya, Lee, Deshpande, Gabriel et~al.}]{parker2022continuous}
Parker, S., Gilstrap, D., Bedoya, A., Lee, P., Deshpande, K., Gabriel, P., et~al. (2022).
\newblock Continuous artificial intelligence video monitoring of icu patient activity for detecting sedation, delirium and agitation.
\newblock In \emph{C35. TOPICS IN CRITICAL CARE AND RESPIRATORY FAILURE} (American Thoracic Society). A5719--A5719
\bibAnnoteFile{parker2022continuous}

\bibitem[{Posch et~al.(2014)Posch, Serrano-Gotarredona, Linares-Barranco, and Delbruck}]{posch2014retinomorphic}
Posch, C., Serrano-Gotarredona, T., Linares-Barranco, B., and Delbruck, T. (2014).
\newblock Retinomorphic event-based vision sensors: bioinspired cameras with spiking output.
\newblock \emph{Proceedings of the IEEE} 102, 1470--1484
\bibAnnoteFile{posch2014retinomorphic}

\bibitem[{Watzlaf et~al.(2010)Watzlaf, Moeini, and Firouzan}]{watzlaf2010voip}
Watzlaf, V.~J., Moeini, S., and Firouzan, P. (2010).
\newblock Voip for telerehabilitation: A risk analysis for privacy, security, and hipaa compliance.
\newblock \emph{International Journal of Telerehabilitation} 2, 3
\bibAnnoteFile{watzlaf2010voip}

\bibitem[{Westbrook et~al.(2011)Westbrook, Duffield, Li, and Creswick}]{westbrook2011much}
Westbrook, J.~I., Duffield, C., Li, L., and Creswick, N.~J. (2011).
\newblock How much time do nurses have for patients? a longitudinal study quantifying hospital nurses' patterns of task time distribution and interactions with health professionals.
\newblock \emph{BMC health services research} 11, 1--12
\bibAnnoteFile{westbrook2011much}

\bibitem[{Wilson et~al.(2020)Wilson, Mart, Cunningham, Shehabi, Girard, MacLullich et~al.}]{wilson2020delirium}
Wilson, J.~E., Mart, M.~F., Cunningham, C., Shehabi, Y., Girard, T.~D., MacLullich, A.~M., et~al. (2020).
\newblock Delirium.
\newblock \emph{Nature Reviews Disease Primers} 6, 90
\bibAnnoteFile{wilson2020delirium}

\end{thebibliography}

\section*{Tables}

\begin{table}[h!]
    \centering
    \caption{\textbf{Performance Metrics of Successive Model Versions for Object Detection.}
    Summary of precision and F1-scores across different versions of the LookDeep Health AI model, highlighting improvements in key tasks as the training data increased. 
    The baseline YOLOv4 model demonstrates initial performance levels, while successive versions (Models v1 to v5) show incremental gains in object detection. 
    With each model iteration, higher precision and F1-scores indicate enhanced detection accuracy and classification reliability, underscoring the impact of additional data and model refinement on real-time patient monitoring capabilities. 
    Evaluation was performed on a fixed dataset containing 10k images.}
    \small
    \renewcommand{\arraystretch}{1.5}
\begin{tabular}{|c|c|c@{\hspace{10pt}}c|}
    \hline
    \textbf{Model Version} & \textbf{Fine-tuning Data Size} & \multicolumn{2}{c|}{\textbf{Object Detection (All)}} \\ \cline{3-4}
    & & \textbf{Precision} & \textbf{F1} \\ \hline
    YOLOv4 (baseline) & \hspace{5pt} n/a \hspace{5pt} & \hspace{5pt} 0.84 \hspace{5pt} & \hspace{5pt} 0.59 \hspace{5pt} \\ \hline
    Model v1 (2022 Q1) & \hspace{5pt} +700 \hspace{5pt} & \hspace{5pt} 0.97 \hspace{5pt} & \hspace{5pt} 0.74 \hspace{5pt} \\ \hline
    Model v2 (2023 Q2) & \hspace{5pt} +2,474 \hspace{5pt} & \hspace{5pt} 0.98 \hspace{5pt} & \hspace{5pt} 0.83 \hspace{5pt} \\ \hline
    Model v3 (2023 Q3) & \hspace{5pt} +10,133 \hspace{5pt} & \hspace{5pt} 0.97 \hspace{5pt} & \hspace{5pt} 0.83 \hspace{5pt} \\ \hline
    Model v4 (2024 Q1) & \hspace{5pt} +28,914 \hspace{5pt} & \hspace{5pt} 0.98 \hspace{5pt} & \hspace{5pt} 0.91 \hspace{5pt} \\ \hline
    Model v5 (2024 Q2) & \hspace{5pt} +34,239 \hspace{5pt} & \hspace{5pt} 0.97 \hspace{5pt} & \hspace{5pt} 0.92 \hspace{5pt} \\ \hline
\end{tabular}
    \label{table:object_detection}
\end{table}

\begin{table}[h!]
    \centering
    \caption{\textbf{Performance Metrics of Successive Model Versions for Object Detection (person), Role Classification, and "Patient Alone" Classification.} Additional results corresponding to Table \ref{table:object_detection} are presented here, focusing on object detection of persons, role classification, and "patient alone" classification tasks.}
    \small
    \renewcommand{\arraystretch}{1.5}
    \begin{tabular}{|c|c@{\hspace{15pt}}c|c|c|}
        \hline
        \textbf{Model Version} & \multicolumn{2}{c|}{\textbf{Object Detection (Person)}} & \textbf{Role Classification} & \textbf{“Patient Alone” Classification} \\ \cline{2-3}
        & \textbf{Precision} & \textbf{F1} & \textbf{(Patient F1)} & \textbf{(F1)} \\ \hline
        YOLOv4 (baseline) & \hspace{5pt} 0.98 \hspace{5pt} & \hspace{5pt} 0.41 \hspace{5pt} & n/a & \hspace{5pt} 0.28 \hspace{5pt} \\ \hline
        Model v1 (2022 Q1) & \hspace{5pt} 0.98 \hspace{5pt} & \hspace{5pt} 0.85 \hspace{5pt} & n/a & \hspace{5pt} 0.86 \hspace{5pt} \\ \hline
        Model v2 (2023 Q2) & \hspace{5pt} 0.97 \hspace{5pt} & \hspace{5pt} 0.89 \hspace{5pt} & n/a & \hspace{5pt} 0.91 \hspace{5pt} \\ \hline
        Model v3 (2023 Q3) & \hspace{5pt} 0.97 \hspace{5pt} & \hspace{5pt} 0.86 \hspace{5pt} & \hspace{5pt} 0.97 \hspace{5pt} & \hspace{5pt} 0.88 \hspace{5pt} \\ \hline
        Model v4 (2024 Q1) & \hspace{5pt} 0.97 \hspace{5pt} & \hspace{5pt} 0.91 \hspace{5pt} & \hspace{5pt} 0.98 \hspace{5pt} & \hspace{5pt} 0.94 \hspace{5pt} \\ \hline
        Model v5 (2024 Q2) & \hspace{5pt} 0.96 \hspace{5pt} & \hspace{5pt} 0.91 \hspace{5pt} & \hspace{5pt} 0.98 \hspace{5pt} & \hspace{5pt} 0.92 \hspace{5pt} \\ \hline
    \end{tabular}
    \label{table:remaining_metrics}
    \end{table}

\begin{table}[ht]
    \centering
    \caption{\textbf{Performance Comparison of Models on Unblurred vs. Face-Blurred Images Across Versions.}
    Evaluation of model performance on unblurred and face-blurred images across different versions. 
    The F1-score measures the model's performance, with the "$\Delta$ F1" column showing the gap between unblurred and face-blurred images. 
    A $\Delta$ value closer to 0 indicates better consistency in model performance between unblurred and face-blurred images.} 
    \small
    \renewcommand{\arraystretch}{1.5}
    \begin{tabular}{|l|c|c|c|c|}
    \hline
    \textbf{Model Version} & \textbf{Evaluation Data Size} & \textbf{Unblurred Images} & \textbf{Face-blurred Images} & \textbf{$\Delta$ F1} \\ 
     & & \textbf{(F1)} & \textbf{(F1)} & \\ \hline
    Model v3 (2023 Q3) & 2,135 & 0.81 & 0.85 & +0.04 \\ \hline
    Model v4 (2024 Q1) & 1,809 & 0.86 & 0.90 & +0.04 \\ \hline
    Model v5 (2024 Q2) & 1,226 & 0.89 & 0.91 & +0.02 \\ \hline
    \end{tabular}
    \label{tab:blurred_vs_unblurred}
    \end{table}    

\section*{Figures}

\end{document}